\pdfoutput=1

\documentclass[11pt]{article}

\usepackage{acl}

\usepackage{times}
\usepackage{latexsym}
\usepackage{graphicx}
\usepackage{amssymb}

\usepackage[T1]{fontenc}

\usepackage[utf8]{inputenc}

\usepackage{microtype}
\usepackage{booktabs}
\usepackage{multirow}
\usepackage{listings}
\lstdefinelanguage{prompt}{
    frame=l,
    framerule=3pt,
    framesep=8pt,
    basicstyle=\small\ttfamily,
    commentstyle=\color{cyan},
    morecomment=[l]{//},
    moredelim=[is][\color{red}\bfseries]{<<<}{>>>},
    moredelim=[is][\color{magenta}\bfseries]{[[[}{]]]},
    moredelim=[is][\color{orange}\bfseries]{===}{===},
    moredelim=[is][\color{olive}\bfseries]{|||}{|||},
}
\lstdefinelanguage{ioexample}{
    frame=shadowbox,
    rulesepcolor=\color{gray},
    framerule=0.5mm,
    rulesep=2mm,
    basicstyle=\small\normalfont,
    commentstyle=\color{cyan},
    morecomment=[l]{//},
    moredelim=[is][\color{red}\bfseries]{<<<}{>>>},
    moredelim=[is][\color{magenta}\bfseries]{[[[}{]]]},
    moredelim=[is][\color{orange}\bfseries]{===}{===},
    moredelim=[is][\color{olive}\bfseries]{|||}{|||},
    moredelim=[is][\bf]{:::}{:::},
    moredelim=[is][\it]{---}{---},
    moredelim=[is][\tt]{+++}{+++},
}
\title{\emph{Previously on} the Stories: Recap Snippet Identification for Story Reading}

\author{Jiangnan Li\textsuperscript{\rm 1,2$\circ$}, Qiujing Wang\textsuperscript{\rm 4}, Liyan Xu\textsuperscript{\rm 3}, Wenjie Pang\textsuperscript{\rm 3}, \\ \bf Mo Yu\textsuperscript{\rm 3$*$}, Zheng Lin\textsuperscript{\rm 1,2}\thanks{\ \ \ Corresponding authors. \textsuperscript{$\circ$} Work done during the Tencent Rhino-bird Research Elite
Program at WeChat.}, Weiping Wang\textsuperscript{\rm 1}, Jie Zhou\textsuperscript{\rm 3}  \\
  \textsuperscript{\rm 1}Institute of Information Engineering, Chinese Academy of Sciences, Beijing, China \\
  \textsuperscript{\rm 2}School of Cyber Security, University of Chinese Academy of Sciences, Beijing, China \\
  \textsuperscript{\rm 3}Pattern Recognition Center, WeChat AI, Tencent Inc. \\
  \textsuperscript{\rm 4}Xi'an Jiaotong University, Xi'an, China \\
  \texttt{\textrm{\{}lijiangnan,linzheng\textrm{\}}@iie.ac.cn,} \texttt{moyumyu@global.tencent.com}
  }

\begin{document}
\maketitle
\begin{abstract}
Similar to the ``previously-on'' scenes in TV shows, recaps can help book reading by recalling the readers' memory about the important elements in previous texts to better understand the ongoing plot.
Despite its usefulness, this application has not been well studied in the NLP community.
We propose the first benchmark on this useful task called Recap Snippet Identification with a hand-crafted evaluation dataset. Our experiments show that the proposed task is challenging to PLMs, LLMs, and proposed methods as the task requires a deep understanding of the plot correlation between snippets. 
\end{abstract}

\section{Introduction}

In TV shows, a clip montage named ``\emph{the previously}'' or ``\emph{previously-on}'' is usually shown at the beginning of an episode, so as to recap to the audiences what happened in previous episodes that can help them understand the current one. 
Likewise, in book reading, a recap of previous plots that are tightly related to the current scene could also assist in reflecting the story's progression in retrospect, attenuating two following challenges in book comprehension.
First, long story books could elicit a significant temporal gap between the current section and its closely related plots in the previous context.
Second, there are parts earlier in the book that did not seem important initially and thus are easy to ignore, which become critical with the developments of new events.
As both cases demand human memory, providing a recap in book reading could offer substantial value in practice.

More formally, we frame it as a research problem to identify \emph{recap snippets}, whereby given the current snippet (a short excerpt from the book), the task objective is to identify snippets in the previous context that contain \emph{related} events or plots to the current one. Since ``\emph{related}'' can be quite subjective among readers, we define explicit criteria to disambiguate the recap process (Section~\ref{definition}, Appendix~\ref{guidelines}), especially with a focus on directly temporal and causal correlations, requiring a deep understanding of the snippet content.



Although researchers~\cite{PlotRetrieval,PERONET,movie_book_align} are studying the story snippets in narratives like novels and films, none of them focuses on the temporal and causal plot associations between snippets and extracts customized recap snippets for the current snippet. AI systems' ability to identify recap snippets still lacks exploration from a scientific view and a good recap snippet identifier also has practical potential for readers to quickly recall the plot when browsing reading APPs and websites. To fill the vacancy, we propose a new task called Recap Snippet Identification. To support it, we present a new dataset dubbed RECINDENT (\textbf{REC}ap snippet \textbf{IDENT}ification) on three novels and two TV productions. RECIDNET for each novel is constituted of: 1) pre-split snippets of the entire book text; 2) human annotation on sampled snippets along with their context history as the test set for evaluation. For each sampled snippet, annotators are asked to label YES/NO on its history snippets up to a certain context window, in regards to whether a certain snippet is plot-related to the current one, according to the defined criteria.

Several baseline approaches are then proposed (Section~\ref{methods}). In particular, with the flourishing of LLMs~\cite{ChatGPT,LLaMA,LLaMA2}, we study whether LLMs can understand the temporal and plot-related associations between snippets in zero-shot and supervised settings as LLMs may have already learned the original books or scripts of our proposed data. 
Besides, due to the lack of training annotation, we designed an auxiliary unsupervised training called Line2Note training, which shows that easily obtained resources like reader notes can be a bridge to connect two snippets with the plot associations. Our empirical results suggest that there still exists a large performance gap between humans and the best baseline, despite the supposedly strong capabilities of LLMs. Overall, this paper is the first to study Recap Snippet Identification with hand-crafted evaluation data and proposed approaches.



\section{Related Work}

\paragraph{Textual Similarity}
AI systems have been studying textual similarity for years, as the similarity between texts can be adapted for a wide range of practices such as information retrieval~\cite{retrieval_survey}, text embedding~\cite{word_embed_survey}, text classifications~\cite{classification_survey}, etc. With the flourishment of pretrained models~\cite{BERT,Roberta}, they are utilized to encode the sentence embedding for further similarity computing. Based on pretrained models, Sentence BERT~\cite{SBERT} is trained in a siamese style; \citet{Bert-flow} transfers the sentence distribution; \citet{whitening} whitens the sentence representations; \citet{SimCSE} and \citet{ConSERT} adapt contrastive learning for better representation transfer; \citet{BGE} propose BGE that is better than E5~\cite{E5} and OpenAI text embedding~\cite{Ada}. For our task, we have a similarity-computing formulation. However, the relationship between a target snippet and its recap requires an understanding of deeper association beyond similarity. Our experiments have demonstrated this.

\paragraph{Narrative Understanding}
There is no prior work on our proposed task. We would like to point out the connection of our work to a wide range of narrative understanding research.

Narrative understanding refers to a kind of complicated process that requires machines to mimic the way humans read to understand the story~\cite{narrative_survey}. For such a purpose, \citet{NarrativeQA} evaluate machines by doing question answering about narratives like novels and scripts. \citet{fairytaleQA} focus on QA about fairy tales. As for understanding the structure of narratives, \cite{TRIPOD} identify turning points of stories. Another way to test if a model can understand narratives is summarization, such as summarizing on books~\cite{BookSum}, screenplays~\cite{screenplay}, and novel chapters~\cite{NovelChapter}. \citet{GNAT,movie_book_align} focus on the alignment between movies and books. Focusing on the detailed plot, \citet{PlotRetrieval} evaluate the retrieval on books to find key plot information. In narratives, characters usually play a crucial role in pushing forward the plot. \citet{LiSCU} study the character identification in summary texts. \citet{TVSHHOWGUESS,yu2022few} ask a model to guess the characters in screenplays based on global memories of the story. \citet{PERONET} identify specific personalities of characters in detailed story snippets. These works cover factors of narratives like events, main plots, characters, and story structures. However, none of them focuses on the plot relationship between story snippets, especially the temporal causality of the same event happening. Therefore, we propose our novel task Recap Snippet Identification for this good. 

\section{Recap and Problem Definitions}\label{definition}
Unlike usual ``previously-on'' in TV shows simply introducing what happened in the previous story, in our setting, recap snippets are defined to have direct plot associations with the target snippet. The direct plot associations refer to that (1) the two snippets have a temporal correlation as they both revolve around the same event; (2) the event of the recap snippet can directly and causally lead to the event in the target snippet (e.g., ``the protagonist leads a protest''$\rightarrow$``the protagonist gets arrested''). 

As the key part of situated story reading, RECIDENT aims to find the recap snippets for a target snippet located in a casual context. These recap snippets are scattered in the whole history context of the target snippet\footnote{Books and TV productions usually use some narrative skills like a flashback, leading to some recap snippets showing in the future of the target snippet. We ignore this kind of recap. }. However, with the history farther from the target, the correlation between them weakens. To reduce overwhelming non-related snippets, for each target snippet, we consider the nearby 60 snippets as recap candidates.

Formally, for a target snippet $\textit{T}_{i}=[s_{(i-w/2)}, ...., s_{(i+w/2)}]$, where $s_i$ is the central sentence of the snippet, $i$ stands for the $i$-th sentence id in the book, and $w$ is the number of sentences in the target snippet. As for the candidate snippets, they are a sequence of consecutive sentences chunked by preset snippet length $w_c$, i.e.,  
$\textit{C}_{i:59}=[s_{({(j-60{w_{c}}+1})}, ..., s_{(j-59{w_{c}})}]$, ..., $\textit{C}_{i:0}=[s_{({j-{w_{c}}+1})}, ..., s_{({j})}]$. There is a small distance gap between $\textit{C}_{i:0}$ and $\textit{T}_{i}$ to prevent direct plot coherency, which will be discussed in the following sections. For each candidate $\textit{C}$, its label $y_{\textit{c}}=1$ if it is a recap snippet, otherwise $y_{\textit{c}}=0$.

\section{RECIDENT Dataset}

\subsection{Dataset Construction}

Story narratives can be in many forms, e.g., long book texts with details and short brief synopses. With the same size of 100-400 words, snippets from books often provide fewer and sparse plots, but snippets from synopses, which are more informative, give more compact and understandable plots. To study the recap snippet identification on story plot in different granularity, we collect data from resources including \textit{classic novels}, \textit{martial chivalry novels} (\textit{Wuxia}\footnote{Wuxia, which literally means "martial heroes", is a genre of Chinese fiction concerning the adventures of martial artists in ancient China. See \url{https://en.wikipedia.org/wiki/Wuxia}.}), \textit{TV series}, and \textit{Anime}\footnote{Anime is hand-drawn and computer-generated animation originating from Japan. We regard it as a kind of TV production. }. As TV series and anime can provide episode synopses, we split these sources of stories into two categories: \textit{Books} and \textit{TV Productions}. We elaborate on the data constructions of the two categories in the following subsections. 

\subsubsection{\textit{Books} Construction}

We collect 2 kinds of books (i.e., classic novels and Wuxia novels). For classic novels, two masterpieces are picked. They are \textit{Notre-Dame de Paris} (NDDP) by Victor Hugo and \textit{The Count of Monte Cristo} (TCOMC) by Alexandre Dumas (père). For Wuxia novels, we pick the most famous \textit{Demi-Gods and Semi-Devils} (DGSD) by Louis Cha. 

\paragraph{Preprocess} As characters are the main meta factors to carry and push the story plot process in the books, we construct target snippets and their candidate recap snippets revolving around characters. For target snippet construction,  we setencize the book into sentences using Spacy, and then use Spacy to recognize the character names in all sentences. By counting the frequencies of characters appearing, we select those characters with appearing times larger than 100 as the main characters. Sentences containing the main characters are filtered out as the central sentences. We further set the number of sentences in target snippets $w$ as 10 for CMC and 7 for the other two. To prevent target snippets from overlapping, we pick up central sentences that can form the longest sequence without overlapping. Until now, the number of central sentences is still large, which can lead to numerous annotating burdens. Therefore, we randomly and evenly sample central sentences so that the target snippets can cover the whole book. Appending $w-1$ sentences around central sentences, target snippets about main characters are produced. The number of target snippets is 187 in NDDP, 400 in TCOMC, and 443 in DGSD. 

With target snippets sealed, their corresponding candidate recap snippets are constructed by sampling the sentence $j$ in $\textit{C}_{i:0}$. The reason why we do not pick up candidates from $i-\frac{w}{2}-1$ is that the most adjacent sentences of the target snippet can directly be regarded as a recap snippet because they can be semantically coherent. To avoid this, we sample the end sentence $j$ of candidates in the range of [$i-10$, $i-20$]. Depending on $j$, we chunk 10 (TCOMC) or 6 sentences as a candidate snippet from back to front. We form 60 candidate snippets for each target. 

\paragraph{Human Annotation} As the recap snippet recognition requires prior knowledge of the story plot in the books, it is hard to find a group of people familiar with these books. Besides, reading the full books is also costly for annotators, which cannot be done in a short period. To this end, we hire two experts in the stories to train annotators. As for famous novels, there are a lot of film and TV versions, which can be useful resources to make annotators quickly understand the plot and characters of the books. For each book, we first train annotators to watch the productions with the least adaption and the corresponding commentary videos, and we then provide detailed plots, character introductions, and specific guidelines for annotators learning. 

 The annotators we hire are all Chinese citizens who have received at least a high school education (The Chinese education system provides lessons about the world's classics). They are required to judge whether a candidate snippet is a recap of the target snippet according to the training that has been done. To make the annotations more convenient, we provide the chapter ID where the target snippet is located along with a brief introduction of the chapter and several former chapters so that annotators can recall what specific plot or events have happened near the target snippet and quickly understand what the big topic the target snippet is in. For each target-candidate pair, we collect 3 answers. The annotation interface is illustrated in the Appendix~\ref{guidelines}. 

 \paragraph{Annotating Quality of \textit{Books}} Before the formal annotation, we train annotators by letting them annotate subsets of target snippets and the average pass rate is $94.71\%$ according to the experts. For the formal annotation, the Fleiss' kappa is 0.8137, which just reaches an almost perfect agreement. 

\paragraph{Chinese-English Alignment} The original book data we collect are in different languages (i.e., English for NDDP and TCOMC, Chinese for DGSD), but the annotation is adapted in Chinese. Therefore, we should get the exact sentence mapping between the two language versions of books so that the position of a snippet can be located in both languages. Following \citet{PERONET}, we utilize Vecalign~\cite{vecalign} and LASER\footnote{\url{https://github.com/facebookresearch/LASER}} to achieve this, which can achieve $>97\%$ match rate according to \citet{PERONET}. 

\subsubsection{\textit{TV Production} Construction}

Different from \textit{Books}, the \textit{TV Production} category does not get snippets from book texts or scripts. We study the recap snippet identification on the synopses of episodes, which summarize the plot in an episode with paragraphs. We collect the episode synopses of TV series \textit{Game of Thrones} (GOT) and Amine \textit{Attack on Titan} (AOT) from their Fandom\footnote{An entertainment site of TV, movies, and games resources for fans. See \url{https://www.fandom.com/}} wiki. The reason for picking these two is that both are high-popularity productions worldwide, so considerable fans are gathering at Fandom and writing high-quality wikis. 

\paragraph{Get Recap By Events} Unlike the snippet in \textit{Books}, we naturally regard paragraphs in an episode synopsis as snippets because they are split by wiki editors acting as an ideal plot unit. All synopsis paragraphs are listed with the episode order to form a long snippet sequence. We then need to construct targets and their candidates from the sequence. By observing that Fandom wikis provide introductions of events that happen in GOT and AOT, we consider utilizing the event information to automatically label recap snippets for target snippets. 

Just like the literal meaning, events in GOT and AOT are specific things happening in the story. For example, GOT has an event called ``Assassination of Robert Baratheon'' and AOT has an event named ``Fall of Wall Maria''. Fandom wiki editors introduce the event by writing its prelude, the event body. Most importantly, for each paragraph or group of sentences, \textbf{an anchor redirecting to which episode it refers to is provided}. To this end, if the paragraph in an event can map to paragraphs in its episode synopsis, we can thread a string of episode paragraphs that highly correlate to a specific event. For an episode paragraph in the thread, the paragraphs before it can act as its recap snippet. 

Following this observation, we use SBERT to map each event paragraph to the episode paragraphs in its corresponding episode. For each event paragraph, SBERT encodes all paragraphs in its episode and the episode paragraphs will be ranked by the similarity score with the event paragraph. If the 3 (or 2) episode paragraphs with the highest score are consecutive paragraphs, they are selected to be the mapping of the event paragraph. If the top 1 paragraph and the top 2 one are not consecutive, only the top 1 episode paragraph is selected. After the mapping, we take all the episode paragraphs mapped in the event body as the target snippets. From the target snippet, we take the last 60 episode paragraphs as candidate snippets. For every candidate snippet, if it is mapped in the event's prelude or in the event body before the mapped position of its target. See Fig.~\ref{event_align} for illustration. 

\paragraph{Quality of \textit{TV Productions}} We hire a big fan who is a story-understanding researcher and is familiar with both productions to test whether the paragraph mapping is good. We sample subsets of events, and the overall mapping rate is $90.36\%$ according to the test. We have tried other mapping tools like Vecalign~\cite{vecalign} and GNAT~\cite{GNAT} and both show bad performance. 

\subsection{Data Statistics}

\begin{table}[]
\centering
\scalebox{0.58}{\begin{tabular}{l|cc|cc|cccc}
\hline
\multirow{2}{*}{} & \multicolumn{2}{c|}{NDDP}         & \multicolumn{2}{c|}{DGSD}          & \multicolumn{2}{c}{TCOMC}                               & GOT                        & AOT  \\ \cline{2-9} 
                  & \multicolumn{1}{c|}{C}     & E    & \multicolumn{1}{c|}{C}     & E     & \multicolumn{1}{c|}{C}     & \multicolumn{1}{c|}{E}     & \multicolumn{1}{c|}{E}     & E    \\ \hline
tgt\_num       & \multicolumn{2}{c|}{159}          & \multicolumn{2}{c|}{397}           & \multicolumn{2}{c|}{307}                                & \multicolumn{1}{c|}{204}   & 194  \\ \hline
tgt\_len       & \multicolumn{1}{c|}{268}   & 153  & \multicolumn{1}{c|}{345}   & 270   & \multicolumn{1}{c|}{418}   & \multicolumn{1}{c|}{237}   & \multicolumn{1}{c|}{187}   & 127  \\ \hline
cand\_len    & \multicolumn{1}{c|}{241}   & 135  & \multicolumn{1}{c|}{290}   & 224   & \multicolumn{1}{c|}{420}   & \multicolumn{1}{c|}{237}   & \multicolumn{1}{c|}{170}   & 111  \\ \hline
hist\_len      & \multicolumn{1}{c|}{17405} & 8273 & \multicolumn{1}{c|}{17668} & 13608 & \multicolumn{1}{c|}{25467} & \multicolumn{1}{c|}{14393} & \multicolumn{1}{c|}{10298} & 6632 \\ \hline
recap\_num        & \multicolumn{2}{c|}{5.6}          & \multicolumn{2}{c|}{7.6}           & \multicolumn{2}{c|}{12}                                 & \multicolumn{1}{c|}{5}     & 11   \\ \hline
\end{tabular}}
\caption{Statistics of RECIDENT. \textit{tgt\_num} is the number of target snippets; \textit{tgt\_num} is the average length of target snippets; \textit{cand\_len} denotes the avg. len. of a single candidate snippet; \textit{hist\_len} denotes the avg. len. of all candidates for a target snippet; \textit{recap\_num} is the avg. num. of recap snippets for a target snippet. }
\label{statistics}
\end{table}

For \textit{Books}, after human annotations, we remove those target snippets without any recap snippets. The statistics of our proposed dataset RECIDENT including \textit{Books} and \textit{TV Productions} are shown in Tab.~\ref{statistics}. In the table, \textit{hist\_len} denotes the average length of the concatenation of all candidate snippets for a target snippet. All the sequence length computing is based on \textsc{TikToken}\footnote{https://github.com/openai/tiktoken} (gpt-3.5-turbo). 

\begin{figure}
    \centering
    \scalebox{0.50}{\includegraphics{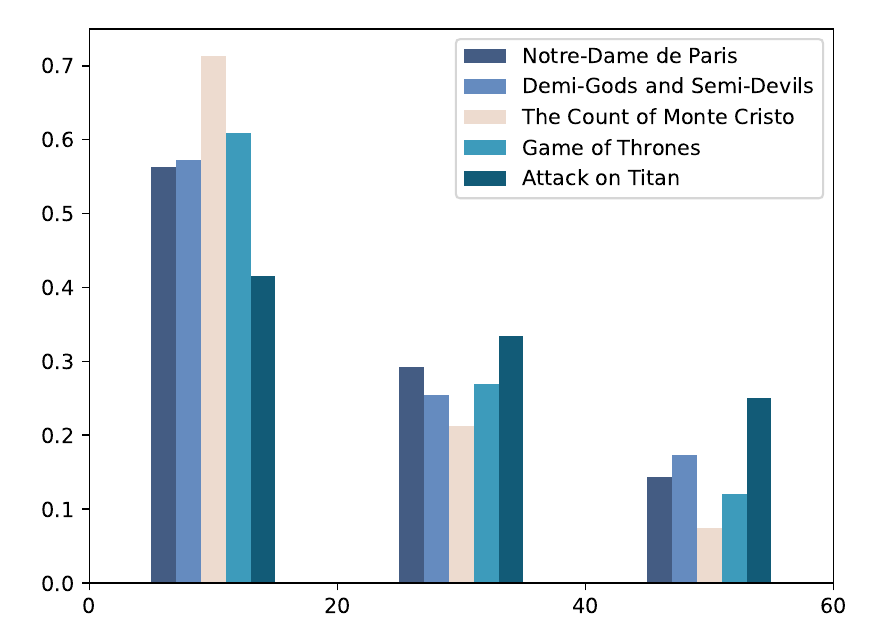}}
    \caption{The proportions of recap snippets in different ranges of the distance between the target and the recap of all datasets. The frequency of recap snippets appearing is in a downtrend with the increasing distance. }
    \label{dis}
\end{figure}

The distribution of recap snippets located at three levels of distance ranges is illustrated in Fig.\ref{dis}. It can be seen that NDDP, DGSD, TCOMC, and GOT have a similar distribution, although TCOMC is most imbalanced as the near snippet is easier to regard as a recap. As for AOT, its recap snippet distribution is the evenest because the main plot in AOT consists of several big events and it is common that several episodes in the same season talk about a story. To this end, the candidate snippets tend to get involved in the same event. In general, all subsets in RECIDENT align with the nature that nearer snippets are more likely to be plot-related. 

\section{Methods}\label{methods}

In this section, we elaborate on the methods for our task. First of all, we claim that characters are the main carriers and prompters of the plot. Therefore, appearing characters can provide some information for recap identification. To this end, we utilize NER tools to recognize characters' names in the target snippet and then filter out candidate snippets in which the recognized characters do not appear. We will adapt this simple filtering (denoted as ``Char-Filter'') to the latter methods when guessing recap snippets. 


\subsection{Prompting Large Language Models}

We leverage the text comprehension capability of LLM to discern recap snippets. In particular, we design two methods, denoted as Listwise and Pairwise prompts.

\textbf{Listwise Prompts} \ \ For each target snippet, we list and input all candidate snippets to LLM in order, and we explain to LLM the intention to extract up to 5 recap snippets providing the most relevant background or prelude information. LLM will follow the instruction to pick up most 5 snippets and explain the reasons. We mainly use ChatGPT~\cite{ChatGPT} as the focused LLM, and the model's max length capacity is 16k. For samples exceeding 15k tokens, we truncate each candidate snippet to the length of 220. The prompt detail can be seen in the Appendix~\ref{prompt_example}. 


\textbf{Pairwise Prompts}\ \ For each candidate snippet of a target snippet, LLM is requested to analyze whether it qualifies as a recap snippet for the target snippet. After identifying all recap snippets for the target snippet, we rank these recaps by the distance between them and the target snippet and select the nearest 5 recap snippets. As a target snippet has 60 recap candidates, getting the results requires 60 requests to ChatGPT, which is costly and time-consuming. Pre-filtering candidates before the Pairwise request can be a feasible way and we will discuss this subsequently. 




\subsection{Unsupervised Line2Note Training}

\begin{figure}
    \centering
    \scalebox{0.26}{\includegraphics{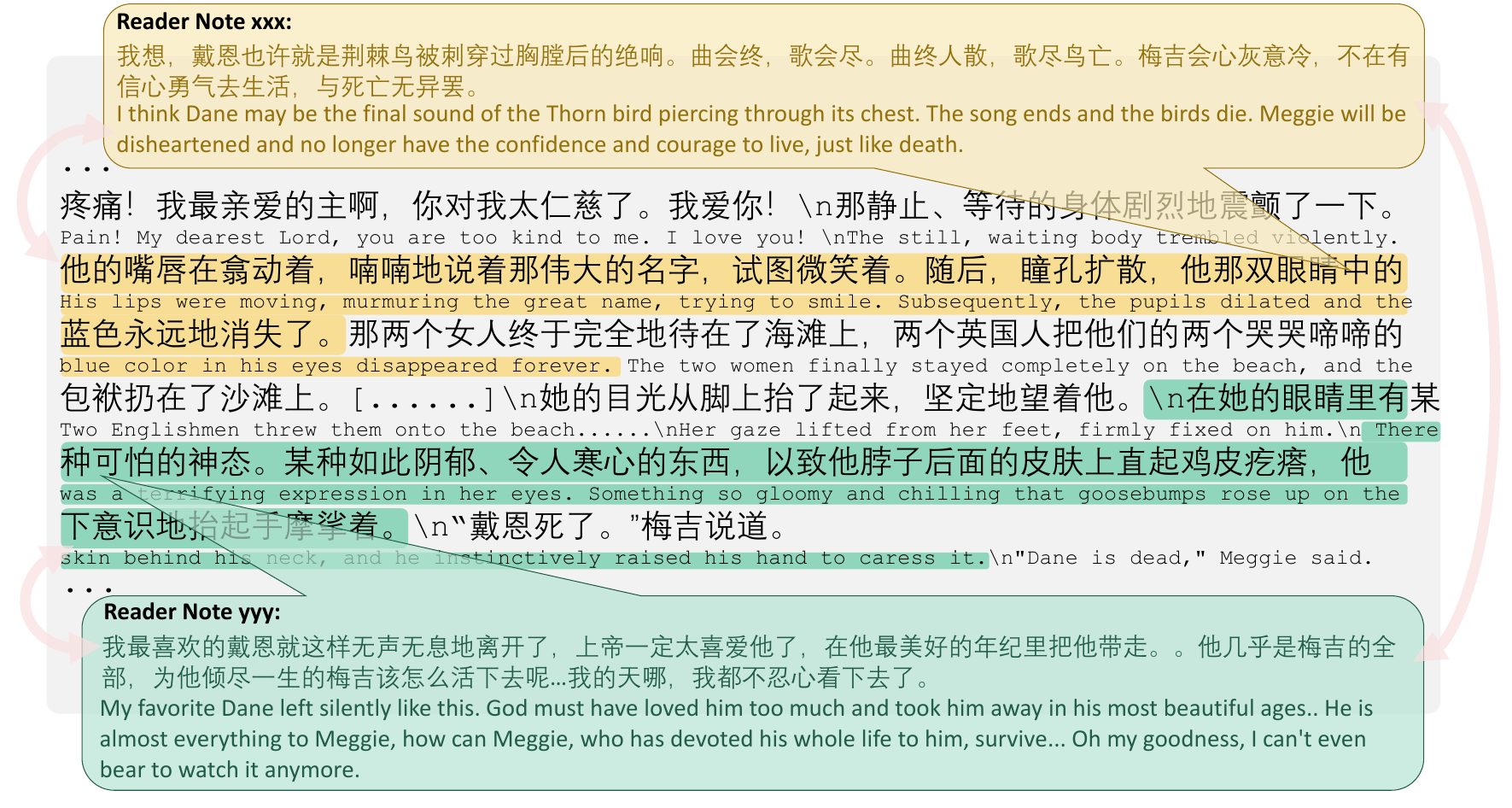}}
    \caption{Reader notes can be the bridge to connect two underlined snippets they attach to. The example shows that both notes comment on the Death of Dane (from The Thorn Birds). }
    \label{l2n_fig}
\end{figure}

\textit{Books are the ladder of human progress. }People often have their own opinions about the plot and characters, and writing down these notes can be a great way for a profound understanding. In reading APPs, readers leave substantial notes on the position they have read to. These notes contain profuse plot-related information. By viewing notes, we find that notes can be a bridge to connect two story snippets if the notes tagged on them talk about a similar plot, as illustrated in Fig.~\ref{l2n_fig}. To this end, we propose a Line2Note Learning method. 

We collect 1623519 notes from 115 classic books of Chinese versions\footnote{It is the number of books, not the number of works because some books are the different translation versions of the same work. } as training data, 131132 notes (96 books) for evaluation, and 140663 (26 books) for testing. We denote a note as $m$ and the span it attaches to as $w_{(s,e)}$, where $s$ and $e$ are the absolute index of the start word and the end word of the span. For short spans, we pad local context around them to supplement them to a specific length. The padded span, named as a \textit{line} by us, is denoted as $w_{(s^{'},e^{'})}$, where $s^{'}\leq s\leq e \leq e^{'}$. We utilize pretrained models, RoBERTa for illustration, to encode the note and the line into embeddings:
\begin{eqnarray}
    \mathcal{M} = RBT(m),\ \ \mathcal{W}=RBT(w_{(s^{'},e^{'})}), \\
    \mathcal{E}_{m} = wsum(Att(\mathcal{M}, 0, l_m, 0, l_m), \mathcal{M}), \\
    \mathcal{E}_{w_{(s,e)}} = wsum(Att(\mathcal{W}, s, e, s^{'}, e^{'}), \mathcal{W}),
\end{eqnarray}
where $RBT$ denotes the encoder;$l_m$ denotes the length of the note; $\mathcal{M}\in \mathbb{R}^{l_m\times d}$ and $\mathcal{W}\in \mathbb{R}^{(e^{'}-s^{'})\times d}$
denote the last hidden states of the encoder; $wsum$ denotes the weighted sum operation of the attention; $\mathcal{E}_{m}\in \mathbb{R}^{1\times d}$ and $\mathcal{E}_{w_{(s,e)}}\in \mathbb{R}^{1\times d}$ are the embeddings for the note and the line. As for, $Att$ is an attention computing the weight for every word:
\begin{eqnarray}
    Att(i,j,k,l,\mathcal{X}) = \mathrm{softmax}(\mathcal{X}\cdot\mathrm{P}\circ \mathrm{Msk}), \\
    \mathrm{Msk}=[\mathrm{F}]*_{(i-k)}+[1]*_{(j-i)}+[\mathrm{F}]*_{(l-j)}, 
\end{eqnarray}
where $\mathrm{P}\in \mathbb{R}^{d\times 1}$ is a trainable parameter; $\mathrm{F}$ denotes $-\infty$; $\mathrm{Msk}$ is the mask covering the word position that is not in the range of $[i,j]$. To this end, for short spans with context padding, the final embedding is the weighted sum of word hidden states in the span. 

For each batch, we sample 20 note-line pairs from the same book. In the 20 pairs, we compute the overlap rate between the spans in every two lines, which is computed by the overlapping length dividing the length of the shorter span. If the overlap rate is larger than 0.8, the two note-line pairs treat each other as a positive sample, otherwise a negative sample. Mind that a span is fully overlapped with itself. Therefore, a note-line pair is always a positive sample to itself. Now, for a line embedding $\mathcal{E}_{w_{(s,e)}}$ with its positive pair set $\mathcal{P}_i$, the training objective:
\begin{equation}
    \mathcal{L}_{w_{(s,e)}} = \sum_{j\in\mathcal{P}_i}{-log(score(\mathcal{E}_{w_{(s,e)}}, \mathcal{E}_{m_{j}}))}, 
\end{equation}
where $score(\mathcal{E}_{w_{(s,e)}}, \mathcal{E}_{m_{j}})$=$\mathrm{linear}([\mathcal{E}_{w_{(s,e)}}\|$ $\mathcal{E}_{m_{j}}\|$ $\mathcal{E}_{w_{(s,e)}}-\mathcal{E}_{m_{j}}\|$ $\mathcal{E}_{w_{(s,e)}}*\mathcal{E}_{m_{j}}])$ from $\mathbb{R}^{4d}$ to $\mathbb{R}^{1}$. 
The training objective can be regarded as 20 times binary classifications for the line to find its positive notes. 

Along the training, lines will realize the contents of the notes attached to them, which are usually related to plot analyses from the view of readers. To this end, if the notes on two lines are talking about the same plot, the embeddings of the two lines will be closer in the feature space than the other unrelated lines. After training, we use the model to compute the snippet embedding $\mathcal{E}$. 


\subsection{Supervised Fine-Tuning on Our Dataset}

Although RECIDENT is proposed to be evaluation data for AI systems' capability of plot retrospect, the number of annotated target-candidate pairs is considerable. To this end, we hold out each subset (e.g., NDDP) for evaluation and construct training pairs from the other four. We supervisedly fine-tune PLMs like RoBERTa and LLMs like LLaMA2~\cite{LLaMA2}. It is worth noting in Tab.~\ref{statistics} that the data is imbalanced as recap snippets are in a long-tailed condition. We adapt the weighted cross-entropy loss to fine-tune PLM. The weight attached to the two classes is computed by $2*\frac{(1/{n_y})^{\alpha}}{\sum_{y^{'}\in (0,1)}{(1/{n_{y^{'}}})^{\alpha}}}$, where $n_y$ is the number of the class and $\alpha\geq 0$ is a scale-controlling factor. As for LLMs fine-tuning, we also tried the over-sampling method to them but we found that they can resist the data imbalance. 

\section{Experiments}

\subsection{Experimental Settings}

\textbf{Evaluation Settings and Metrics}\ \ We adopt two settings. One standard-setting allows the methods to freely select what they think are recap snippets. The other requires the methods to select 5 recap snippets they believe are most relevant. The latter mimics the realistic application scenario where human readers have limited bandwidth to glance at the recommended snippets during reading. The reason why selecting 5 is that it is a proper bandwidth and more importantly performance behaviors of all methods follow the same trend when setting to 10 for AOT and TCOMC. For the first setting, we report Recall, Precision, and F1 score, which will be shown in the Appendix~\ref{free_f1}. As for the second setting, we use Recall@5 and Precision@5 scores, and their F1 score. The Recall@5 is computed by averaging all Recall@5 scores of target snippets, and so does the Precision@5. As for the F1 score, it is computed based on the Recall@5 and Precision@5 scores. 

\begin{table}[]
\centering
\scalebox{0.45}{
\begin{tabular}{@{}l|ccc|ccc|ccc|c@{}}
\toprule\toprule
\multirow{2}{*}{}       & \multicolumn{3}{c|}{NDDP} & \multicolumn{3}{c|}{DGSD}  & \multicolumn{3}{c|}{TCOMC} & \multicolumn{1}{c}{Avg.} \\ \cmidrule(l){2-11} 
                        & R@5          & P@5            & F1@5      & R@5    & P@5          & F1@5     & R@5    & P@5          & F1@5 & F1@5\\ \midrule\midrule
\multicolumn{11}{c}{\textit{zero-shot}} \\
Closest 5 Snippets          & 20.82        & 19.64          & 20.21    &  25.34  &  31.59       & 28.12   &  33.36~~~ &  67.49       & 44.66 & 31.00 \\
RoBERTa                 & 21.03        & 21.18          & 21.10    &  22.73  &  31.49       & 24.60   & 18.32  &  39.48       & 25.03 & 23.58\\
SBERT                   & 21.90        & 23.67          & 22.75    & 23.38   &  33.25       & 27.45   & 20.70  &  44.56       & 28.27 & 26.16\\
\quad  + Char-Filter     & 26.06        & 27.51          & 26.77    & 25.26   &  34.77       & 29.26   & 24.16  &  50.20       & 32.62 & 29.55\\
ChatGPT Listwise     & 7.38         & 8.88           & 8.06     &  6.37      & 9.22    & 7.53       & 3.28   &  8.66        & 4.76  & 6.78\\
\quad  + Char-Filter     & 18.54        & 20.02          & 19.25    &  10.24     & 11.75   & 10.94      & 13.01~~~  &  27.63       & 17.69 & 15.96\\
\midrule
\multicolumn{11}{c}{\textit{fine-tune -- supervised}} \\
RoBERTa Pw Rank         &  23.43       &   26.06        &  24.67   &  23.24  &   31.95      & 26.91   &  27.29 &  56.47       & 36.80 & 29.46\\
SBERT Pw Rank           &  21.29       &   21.27        &  21.33   &  23.70  &   30.74      & 26.77   &  26.97 &  55.84       & 36.37 & 28.16\\
InternLM2 Pw Rank       & 30.46        &   26.98        &  28.61   &  26.20  &   42.93      & 32.54   &  32.62 &  \textbf{71.94}       & \textbf{44.89} & 35.35\\
\quad  + Char-Filter    & 28.64        &   31.13        &  29.84   &  25.59  &   \underline{43.05}      & 32.10   &  31.40~~~ &  \textbf{71.95}       & \underline{43.72} & 35.22\\
\midrule
\multicolumn{11}{c}{\textit{fine-tuning -- unsupervised}} \\
Roberta l2n             & 30.24        & 29.35          & 29.79    & \underline{29.18}   & 38.99        & \underline{33.38}   &  26.76~~~ &  55.18       & 36.04 & 33.07\\
SBERT l2n               & 29.42        & 29.47          & 29.44    & 26.82   & 36.83        & 31.04   &  28.89 &  58.44       & 38.67 & 33.05\\
\midrule
\midrule
\multicolumn{11}{c}{\textit{pipeline systems (Line2Note + ChatGPT Pw$^\dag$)}} \\
 Our full system      & \underline{30.42}        & \textbf{32.93}          & \textbf{31.63}    &  \textbf{29.55}     & 40.47   & \textbf{34.16}      & \textbf{37.17}  &  49.74       & 42.54 & \textbf{36.11}\\
 \quad  w/ InternLM2-Pw  & \textbf{30.69} & \underline{31.27} & \underline{30.98} & 25.14 & \textbf{44.22} & 32.06 & \underline{32.62} & \textbf{71.94} & \textbf{44.89} & \underline{35.98} \\
\midrule
Human Performance$^{*}$ & 43.65        & 54.00          & 48.28    & 52.94   & 69.55        & 60.12   &  44.32 &  82.95       & 57.77 & 55.39\\
\bottomrule\bottomrule
\end{tabular}
}
\caption{Recall@5, Precision@5, F1 scores of methods on the Chinese evaluation data of RECIDENT. $^*$ stands for a subset of the evaluation data that is predicted by humans who have never read the book. Pw is the abbreviation of \textit{pairwise}. The bold result is the best and the underlined one is the second best except \textit{Closest 5} and \textit{Human}. }
\label{chinese_main}
\end{table}

\textbf{Baselines}\ \ We split baselines into 2 categories: \textit{zero-shot}, \textit{fine-tune}. In \textit{zero-shot} style, \textbf{RoBERTa} and \textbf{SBERT} are selected. We study the performance of LLM by utilizing \textbf{ChatGPT Listwise}. Besides, due to the feature of RECIDENT, we also care about the outcome if the closest 5 snippets are chosen, and we name this simple method as \textbf{Closest 5 Snippets}. In \textit{fine-tune} style, we supervisedly fine-tune \textbf{RoBERTa Pairwise}, \textbf{SBERT Pairwise}. For the English data, we also fine-tune LoRA~\cite{LoRA} on Llama2, which is denoted as \textbf{Llama2 Pairwise}. For Chinese data, we fine-tune LoRA on InternLM2~\cite{InternLM}, denoting \textbf{InternLM2 Pairwise}. Besides, the unsupervised Line2Note training is adapted on RoBERTa and SBERT, which can be denoted as \textbf{RoBERTa l2n} and \textbf{SBERT l2n}. As mentioned ChatGPT Pairwise requires 60 requests to get the results, we reduce the burden by adapting ChatGPT Pairwise only on the top-15 snippets filtered by l2n models, which is denoted as our \textbf{full pipeline system}. 

\textbf{Implementation}\ \ Depending on the length of the prompt, we use \textit{gpt-3.5-turbo} and \textit{gpt-3.5-turbo-16k} for ChatGPT prompting. For English evaluation data, \textit{RoBERTa-base}, \textit{all-mpnet-base-v2}, and \textit{llama-2-7b-hf} are adapted. For Chinese evaluation data, \textit{chinese-roberta-wwm-ext-base}, \textit{paraphrase-multilingual-mpnet-base-v2}, and \textit{interlm2} are utilized. The results of models unable to directly compute rank scores are ranked by the distance from the target snippet. As for other hyper-parameters, see the Appendix~\ref{implementation_details}. 

\subsection{Main Results}

\begin{table*}[]
\centering
\scalebox{0.6}{
\begin{tabular}{@{}l|ccc|ccc|ccc|ccc|ccc|c@{}}
\toprule\toprule
\multirow{2}{*}{}       & \multicolumn{3}{c|}{Notre-Dame de Paris} & \multicolumn{3}{c|}{Demi-Gods \& Semi-Devils}  & \multicolumn{3}{c|}{The Count of Monte Cristo}  & \multicolumn{3}{c|}{Game of Thrones} & \multicolumn{3}{c|}{Attack on Titan}& \multicolumn{1}{c}{Avg.}\\ \cmidrule(l){2-17} 
                             & R@5     & P@5    & F1@5    &  R@5~~  &  P@5    & F1@5    & R@5    & P@5    & F1@5   & R@5    & P@5     & F1@5   & R@5    & P@5    & F1@5  & F1@5 \\ \midrule\midrule
\multicolumn{17}{c}{\textit{zero-shot}} \\ 
Closest 5 Snippets               & 20.82   & 19.64  & 20.21   &  25.34~~&  31.59  & 28.12   & 33.36  & 67.49  & 44.66  & 35.45  &  30.98  & 33.06  & 19.55  & 27.84  & 22.97 & 29.80 \\
RoBERTa                      & 15.00   & 15.62  & 15.30   &  16.66~~&  24.28  & 19.76   & 12.62  &  39.48 & 17.84  & 18.29  &  17.06  & 17.65  & 16.64  & 32.16  & 21.93 & 18.50 \\
SBERT                        & 25.26   & 25.80  & 25.53   &  26.67~~&  35.06  & 30.29   &  26.81 & 54.40  & 35.92  & 47.56  &  36.08  & 41.03  & \textbf{25.41}  & \textbf{42.99}  & \textbf{31.94} & 32.94 \\
\quad  + Char-Filter          & 25.30   & 26.25  & 25.77   &  \underline{27.29}~~&  35.26  & 30.77   &  27.37 & 56.12  & 36.79  & \textbf{48.54}  &  36.16  & 41.45  & \underline{23.15}  & \underline{42.81}  & \underline{30.05} & 32.97 \\
ChatGPT Listwise         & 7.55    & 9.70   & 8.49    &  6.00~~ &  9.35  & 7.31    &  3.02  &  7.89  & 4.37   & 26.15  &  24.61  & 25.37  & 15.77  & 30.52  & 20.80 & 13.67 \\
\quad  + Char-Filter          & 22.97   & 22.22  & 22.59   &  9.92~~ &  12.71 & 11.14    &  15.19  &  29.87  & 20.14   & 38.87  &  31.94  & 35.07  & 17.02  & 34.48  & 22.79 & 22.35 \\
\midrule\multicolumn{17}{c}{\textit{fine-tune}} \\
Roberta Pw Rank             &  18.40       &   22.24        &  20.14   &  19.06~~  &   28.59      & 22.87 & 17.20 &  42.86  & 24.55 &  29.48    &  26.96       & 28.16 &  17.69    &  36.39       & 23.80 & 23.90 \\
SBERT Pw Rank               &  21.14       &   22.63        &  21.86   &  23.62~~  &   30.63      & 26.67 & 21.36 &  48.20  & 29.60 &  28.40    &  26.39       & 27.36 &  17.13    &  35.70       & 23.15 & 25.73 \\
Llama2Lora Pw Rank          &  \underline{26.05}       &   \textbf{32.50}        &  \underline{28.92}   &  27.15~~  &   \textbf{39.38}      & \underline{32.14} & \underline{28.13}&  \textbf{68.39}   & \underline{39.86} &  \underline{44.92}    &  \textbf{45.67}      & \textbf{45.29} &  21.61   &  36.29      & 27.09 & \textbf{34.62}\\
\quad  + Char-Filter        &  {23.81}       &   {32.18}        &  {27.37}   &  26.82~~  &   39.38      &{31.91} & \underline{28.13}&  \textbf{68.39}   & \underline{39.86} &  {42.69}    &  {44.60}      & {43.63} &  21.05   &  35.23      & 26.36 & 33.83\\
\midrule
\multicolumn{17}{c}{\textit{pipeline systems (SBERT (Char-Filter) + ChatGPT Pw)}} \\
Our full system      & \textbf{26.59}   & \underline{32.38}  & \textbf{29.11}   & \textbf{28.49}~~ & \underline{38.43}  &  \textbf{32.72}   & \textbf{29.13}   & \underline{63.98}  &  \textbf{40.03}  & 40.34  &  \underline{45.15}  &  \underline{42.61}  & 22.33 & 34.49  & 27.10 & \underline{34.31} \\
\bottomrule\bottomrule
\end{tabular}
}
\caption{Recall@5, Precision@5, F1@5 scores of different methods on the English evaluation data of RECIDENT. Pw is the abbreviation of \textit{pairwise}. }
\label{english_main}
\end{table*}

The results of models on the English evaluation data can be viewed in Tab.~\ref{english_main} and the Chinese counterparts are illustrated in Tab.~\ref{chinese_main}. Seen from both tables, the models show similar behaviors in English and Chinese data. The closest 5 snippets achieve good performance to confirm the nature that nearer snippets are more plot-related. However, it does not mean that the nearest snippets are always the best for readers as they possibly have just read the nearest snippets minutes ago. Other snippets should have the chance to be selected. As for RoBERTa and SBERT, RoBERTa's performance demonstrates that it is not a good similarity scorer as SBERT. However, after supervised fine-tuning, RoBERTa can be a better pairwise recap identifier than before while SBERT hurts its performance. The reason for this may be that SBERT is an already well-adapted sentence similarity scorer and the pairwise recap fine-tuning can hurt its original ability as recap identification is more than just similarity computing. On the contrary, with our proposed unsupervised Line2Note training, both RoBERTa and SBERT receive significant performance gains against original and supervised fine-tuning ones, seen from the average F1 scores in Tab.~\ref{chinese_main}. We attribute it to that models learn the plot association between snippets by the bridge built by topic- or event-relevant notes. 

As for the comprehension capability of LLM to discern recap snippets, ChatGPT, LLaMA2, and InternLM2 are tested. ChatGPT using Listwise Prompts shows the worst performance at @5 metrics on both languages among all methods. The reason for this may be that giving all the history at once can bring in copious noises interfering with ChatGPT. We have formerly mentioned that characters can be pre-filtering information. To this end, we utilize Char-Filter on ChatGPT Listwise. ChatGPT Listwise equipped with Char-Filter has lifted performance in both languages, which shows that characters showing off can help models directly ignore a lot of noise snippets. However, the performance of ChatGPT Listwise is still unsatisfactory, which may indicate that ChatGPT is not a good ranker for recap identification especially when the input context is lengthy. We also utilize Char-Filter to SBERT and Char-Filter can reduce some noises for it. For open-source LLMs, LLaMA2 and InternLM2 are activated by pairwise fine-tuning as both of them achieve almost the best performance against RoBERTa, SBET, and ChatGPT Listwise. When utilizing Char-Filter to them, no gain is shown and the performance is even hurt. We think the reason may be that LLMs own NER ability to some extent and they can use it when guessing the recap in the pairwise style, but NER tools may make mistakes (e.g., wrong extracted names) and cannot figure out a character's alternative names (e.g., \textit{Abbé Busoni} is \textit{Edmond Dantès}). Based on these observations, LoRA fine-tuning LLMs with pairwise recap data can be a possible solution to our task. Nonetheless, more annotated data, which are costly to access, are required to fine-tune LLMs. As reader notes are easier to collect, in future work, we consider fine-tuning LLMs with our Line2Note training. 

\begin{table}[]
\centering
\resizebox{\columnwidth}{!}{
\begin{tabular}{l|ccc|ccc}
\toprule
\multirow{2}{*}{Method}                      & \multicolumn{3}{c|}{NDDP} & \multicolumn{3}{c}{DGSD}  \\ 
\cline{2-7} 
 & R@5  & P@5  & F1@5  & R@5  & P@5  & F1@5 \\  \midrule
pairwise (all 60)       & 29.53 & 28.39 & 28.95 &28.06 &36.17 &31.60  \\
pairwise (l2n top15)         & 30.42 & 32.93 & 31.63  &29.55& 40.47&34.16  \\
\midrule
listwise & 18.54        & 20.02          & 19.25    &  10.24~~     & 11.75   & 10.94 \\
\bottomrule
\end{tabular}
}
\caption{Comparison of ChatGPT Pairwise approach with and without filtering candidate snippets on Chinese NDDP and DGSD. }
\label{chatgpt-prompts}
\end{table}

\begin{table}[]
\centering
\resizebox{\columnwidth}{!}{
\begin{tabular}{l|ccc|ccc}
\toprule
\multirow{2}{*}{Method}                      & \multicolumn{3}{c|}{GOT} & \multicolumn{3}{c}{AOT}  \\ 
\cline{2-7} 
 & R@5  & P@5  & F1@5  & R@5  & P@5  & F1@5 \\  \midrule
SBERT       & 48.44 & 37.07 & 42.00 & 25.33 & 43.28 & 31.96  \\
\quad  w/o event names         & 42.50 & 31.73 &  36.33 & 22.25 & 38.14 & 28.10 \\
\bottomrule
\end{tabular}
}
\caption{The performance of SBERT without event names will decay. }
\label{wo_event}
\end{table}

\paragraph{ChatGPT with Pairwise Prompts}
Besides Listwise Prompts, ChatGPT can also respond with Pairwise Prompts to achieve good performance as other models. However, it suffers from expense and response time. Therefore, we propose a pipeline system: small SBERT and l2n models first rank the top 15 snippets, and ChatGPT then utilizes Pairwise Prompts on them. The results show that the full system can be the best against other fine-tuned models. To show whether our pipeline system hurts the original performance of ChatGPT, we use ChatGPT Pairwise to process all snippets for NDDP and DSGD in Chinese. The results in Tab.~\ref{chatgpt-prompts} show that l2n models firstly filtering snippets can even bring performance gains. We also replace ChatGPT Pairwise with InternLM2 Pairwise and the performance does not drop. To this end, the pipeline can be a simple but effective system to reduce burdens. 

\begin{figure}
    \centering
    \scalebox{0.25}{
    \includegraphics{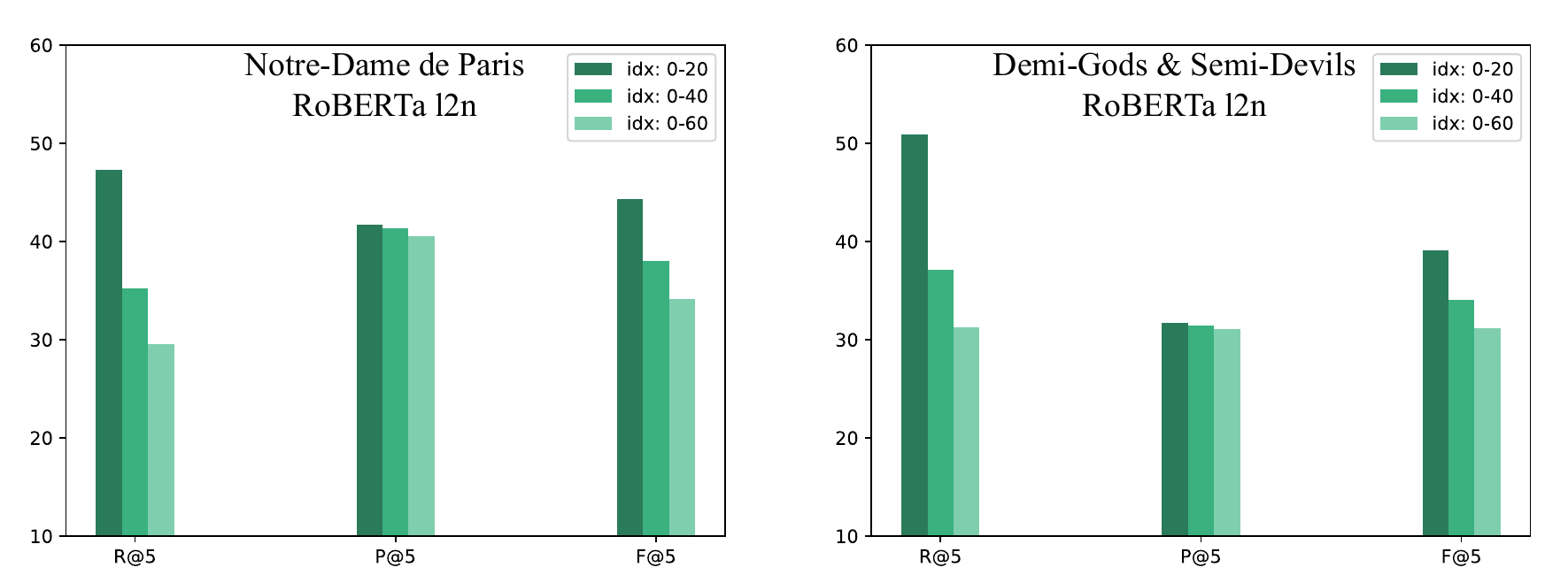}
    }
    \caption{Performance of l2n when only considering the nearest 20/40/60 snippets on NDDP and DGSD. As the distance increases, the identification becomes harder. }
    \label{dis_performance_main}
\end{figure}

\paragraph{The Effect of Distance}
As illustrated in Fig.~\ref{dis_performance_main}, when only identifying recap snippets from the nearest 20 snippets, the l2n model performs the best results. With more snippets considered, all metrics drop showing that the identification becomes more difficult (More illustrations on other subsets can be viewed in Fig.~\ref{dis_performance_appendix}). In most cases, Recall@5 shows a more drastic decrease than Precision@5. The reason is that models tend to rank those near snippets at the top as recaps, and for those farther recaps, it is harder for models to recognize them. This phenomenon conforms to the nature that local context is more plot-related. However, it does not mean that farther recaps are not that crucial because narrative skills like ``foreshadowing'' can make readers more surprised to stimulate their interest. This can raise a more interesting direction that we truly rank all annotated recaps so that their priorities can be considered. However, it can be a more challenging and opinion-divergent task for annotators. 

\paragraph{The Effect of Event Names}
As all models on GOT and AOT use the name of the event at which the target snippet is, we remove the event name to show that the event name can help models recognize recaps. As shown in Tab.~\ref{wo_event}, SBERT removing the event name from the target snippet performs worse than that with event names. Intuitively, snippets involved in an event are more related. Resources from Fandom can provide such information, but it is impractical that snippets in real book reading get annotated with their events, especially when the snippet is too short for machines to summarize. Further study of the event in snippets can be a future work for recap identification. 

\section{Conclusion}

To verify AI systems' ability to capture and analyze the temporal and plot-related associations between story snippets, we propose a new task of identifying recap snippets for a target snippet for story reading. To support the task, we present a new dataset RECIDENT including book and TV productions. Experiments show that the task is challenging in overcoming the long-text noises and understanding the plot correlation between snippets in many factors. 

\bibliography{anthology,custom}

\newpage
\appendix

\section{Example ChatGPT Prompts}\label{prompt_example}
\label{sec:chatgpt-prompts}
\paragraph{Listwise Prompts}
The Listwise Prompt guessing top 5 recap snippets is shown in Fig.~\ref{fig:prompt-list-top5-tv}, and Listwise Prompt freely guessing recap snippets is provided in Fig.~\ref{fig:prompt-list-all-tv}. TV\_PRODUCTION\_TYPE is set to ``TV shows'' for GOT and ``Animes'' for AOT. 
\begin{figure*}[t!]
    \centering
\begin{lstlisting}[language=prompt]
<<<[SYSTEM]>>>
You are an expert on reading and understanding [[[TV_PRODUCTION_TYPE]]] stories. You will be performing certain analysis on given story snippets.

<<<[USER]>>>
You are an expert on reading and analyzing a wide variety of stories of "[[[TV_PRODUCTION_TYPE]]]". Give you a target story snippet related to the event "[[[EVENT_NAME]]]" from "[[[TV_PRODUCTION_NAME]]]". Suppose you are reading the target story snippet as follows:
|||[Target snippet]|||
[[[{{target_snippet}}]]]

You are hoping to understand its important background or prelude information. Please determine which of the following previous snippets from the same story provide the most relevant background or prelude information as a recap, where each previous snippet is identified by a number mark:
|||[Previous snippets]|||
(1) [[[candidate snippet 1]]]
(2) [[[candidate snippet 2]]]
...
(n) [[[candidate snippet n]]]

You can pick up to 5 snippets from the previous snippets as recaps revolving around the event "[[[EVENT_NAME]]]" for the target snippet. For each picked recap snippet, per line, you should firstly guess the number mark of it and then provide your explanation of why you select it as a recap in the format of ({number}) - {explanation}. Please do not copy the whole original text of the picked snippet.
\end{lstlisting}
    \caption{Listwise top5 guessing prompt for \textit{TV Productions}.}
    \label{fig:prompt-list-top5-tv}
\end{figure*}

\begin{figure*}[t!]
    \centering
\begin{lstlisting}[language=prompt]
<<<[SYSTEM]>>>
You are an expert on reading and understanding [[[TV_PRODUCTION_TYPE]]] stories. You will be performing certain analysis on given story snippets.

<<<[USER]>>>
You are an expert on reading and analyzing a wide variety of stories of "[[[TV_PRODUCTION_TYPE]]]". Give you a target story snippet related to the event "[[[EVENT_NAME]]]" from "[[[TV_SHOW_NAME]]]". Suppose you are reading the target story snippet as follows:
|||[Target snippet]|||
[[[{{target_snippet}}]]]

You are hoping to understand its important background or prelude information. Please determine which of the following previous snippets from the same story provide the relevant background or prelude information as a recap, where each previous snippet is identified by a number mark:
|||[Previous snippets]|||
(1) [[[candidate snippet 1]]]
(2) [[[candidate snippet 2]]]
...
(n) [[[candidate snippet n]]]

Your goal is to pick up all the snippets that can act as recaps for the target snippet from previous snippets, revolving around the event "[[[EVENT_NAME]]]" for the target snippet. For each picked recap snippet, per line, you should firstly guess the number mark of it and then provide your brief explanation of why you select it as a recap in the format of ({number}) - {explanation}. Please do not copy the whole original text of the picked snippet.
\end{lstlisting}
    \caption{Listwise freely guessing prompt for \textit{TV Productions}.}
    \label{fig:prompt-list-all-tv}
\end{figure*}

\begin{figure*}[t!]
    \centering
\begin{lstlisting}[language=prompt]
<<<[SYSTEM]>>>
You are an expert on reading and understanding book stories. You will be performing certain analysis on given story snippets.

<<<[USER]>>>
You are an expert on reading and analyzing a wide variety of stories of fictions. Give you a target story snippet from "[[[BOOK_NAME]]]". Suppose you are reading the target story snippet as follows:
|||[Target snippet]|||
[[[{{target snippet}}]]]

You are hoping to understand its important background or prelude information. Please determine which of the following previous snippets from the same story provide the most relevant background or prelude information as a recap, where each previous snippet is identified by a number mark:
|||[Previous snippets]|||
(1) [[[candidate snippet 1]]]
(2) [[[candidate snippet 2]]]
...
(n) [[[candidate snippet n]]]
You can pick up to 5 snippets from the previous snippets as recaps for the target snippet. For each picked recap snippet, per line, you should firstly guess the number mark of it and then provide your explanation of why you select it as a recap in the format of ({number}) - {explanation}. Please do not copy the whole original text of the picked snippet.
\end{lstlisting}
    \caption{Listwise top5 guessing prompt for \textit{books}.}
    \label{fig:prompt-list-top5-book}
\end{figure*}

\begin{figure*}[t!]
    \centering
\begin{lstlisting}[language=prompt]
<<<[SYSTEM]>>>
You are an expert on reading and understanding book stories. You will be performing certain analysis on given story snippets.

<<<[USER]>>>

You are an expert on reading and analyzing a wide variety of stories of fictions. Give you a target story snippet from [[[BOOK_NAME]]]. Suppose you are reading the target story snippet as follows:
|||[Target snippet]|||
[[[{{target snippet}}]]]

You are hoping to understand its important background or prelude information. Please determine which of the following previous snippets from the same story provide the relevant background or prelude information as a recap, where each previous snippet is identified by a number mark:
|||[Previous snippets]|||
(1) [[[candidate snippet 1]]]
(2) [[[candidate snippet 2]]]
...
(n) [[[candidate snippet n]]]
Your goal is to pick up all the snippets that can act as recaps for the target snippet. For each picked recap snippet, per line, you should firstly guess the number mark of it and then provide your brief explanation of why you select it as a recap in the format of ({number}) - {explanation}. Please do not copy the whole original text of the picked snippet.
\end{lstlisting}
    \caption{Listwise freely guessing prompt for \textit{books}.}
    \label{fig:prompt-list-all-book}
\end{figure*}

\paragraph{Pairwise Prompts}
The pairwise prompt used by ChatGPT is provided in Fig.~\ref{fig:prompt-pair}.
\begin{figure*}[t!]
    \centering
\begin{lstlisting}[language=prompt]
<<<[SYSTEM]>>>
Please assess whether the content of the candidate segment helps you understand the content of the target segment.

<<<[USER]>>>
Please determine whether the candidate segment contributes to the understanding and reading of the target segment. Candidate segments are generally helpful to readers in understanding the target segment when there is a strong connection between them, especially if there is an event or causal relationship between them. First, I will provide you with the candidate segment, and then I will give you the target segment. Please assess whether the content of the candidate segment helps you read and understand the target segment.

Please read the following candidate segment from the former context :
|||[Candidate segment]|||
[[[{{ candidate segment }}]]]

Please read the following target segment:
|||[Target segment]|||
[[[{{ target segment }}]]]


Please assess whether the content of the candidate segment helps you understand the content of the target segment. Please analyze and explain your determination first, then answer Yes/No at the end. The response format should be in the format like: "Explanation
Yes/No".
\end{lstlisting}
    \caption{Pairwise prompt.}
    \label{fig:prompt-pair}
\end{figure*}
\section{Implementation Details}\label{implementation_details}

For supervised fine-tuning, we train RoBERTa and SBERT for 10 epochs with the learning rate from [1e-5, 3e-5]; batch size of 32; the max length of the snippet from [128, 256]; the linear decay with a warmup rate of 0.1. The alpha factor in the weighted loss is set to 0 for RoBERTa and set to 1 for SBERT (SBERT is highly effected by the data imbalance but RoBERTa can resist it). For LLaMA2 and InternLM2, we tried the standard sampler and weighted over-sampler with the weight computed by the same formula as that used in weighted loss, and we found that LLMs can also resist data imbalance. We use a standard sampler for these LLMs. For LoRA configuration, we set $r$ to 8, set $\alpha$ to 16, and add LoRA to weights in ``qkv'' with the dropout rate of 0.05. The learning rate for LLMs is set to 3e-4. 

For Line2Note training, the training batch size is set to 20 and the evaluating batch size is set to 12. The learning rate is searched from [3e-5, 2e-5, 1e-5, 5e-6, 1e-6]. The max length of the snippet is set to 256. The Line2Note models are trained with 2 epochs. 

\section{Annotation Notes and Interface}\label{guidelines}

We first explain to annotators such a scenario: A reader continues their reading after a long period of time and forgets what has happened before the checkpoint they left a bookmark to. The reader does not want to read the whole previous context again. Please help the reader find recap snippets that are highly plot-related and plot-causal to the snippet where the checkpoint is (target snippet). By reading the recap snippets, the reader can quickly catch up and recall the plot so that they can conveniently continue reading. 

According to annotators, some snippets are kind of plot-related to the target snippet by several reasoning hops. For example, in \textit{The Count of Monte Cristo}, the target snippet is about Edmond Datens getting arrested due to a letter of accusation, where there are two candidate snippets: one is about Danglar jealous about Edmond Datens being selected to the next captain, and the other is about Danglar conspiring with Fernand about framing Edmond. There is an event chain of ``Danglar jealous''$\rightarrow$``Danglar framing Edmond''$\rightarrow$``Edmond arrested''. By two hops, ``Danglar jealous'' can be a recap snippet to ``Edmond arrested'' and such cases frequently appear in the context, leading to dozens of recaps. With over number of recaps, recaps will lose their meaning of helping readers quickly recall the plot because they have to read a large range of context. Besides, too many reasoning hops will make annotators hardly reach an agreement as people have different standards of the proper number of reasoning hops. To avoid our annotation becoming opinion-divergent and redundant, we confirm that annotators should only pick up the direct recaps with only one reasoning hop (``Danglar framing Edmond'' directly results in ``Edmond arrested''). 

\begin{figure*}
    \centering
    \scalebox{0.55}{
    \includegraphics{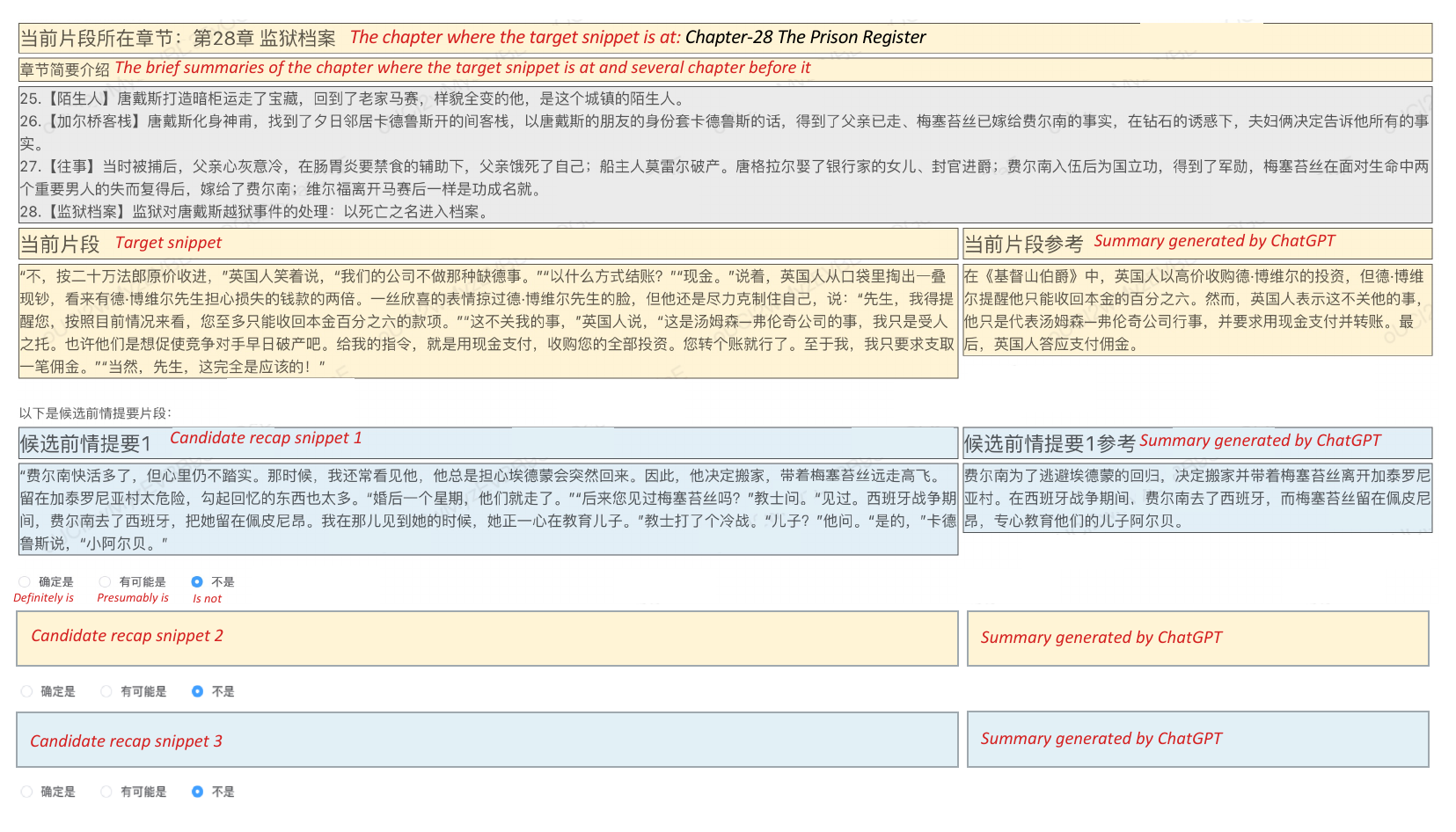}
    }
    \caption{The annotating interface for annotators.}
    \label{annotating_interface}
\end{figure*}

For each target snippet, if only one candidate snippet is provided to an annotator each time, the annotator needs to repeatedly read a target snippet 60 times. If giving all candidate snippets to the annotator, it will be too boring and burdensome for them. Therefore, each time, we provide 3 candidate snippets to the annotator for a target snippet. The annotating interface is shown in Fig.~\ref{annotating_interface}. In the interface, we additionally provide the chapter where the target snippet is and the brief summaries of the chapter and several chapters before it. The information can help annotators locate the snippet and quickly realize the plot. To further reduce the difficulty of the annotation, we use ChatGPT to summarize the content of the snippet. These summaries can be useful for annotators if the snippet is too abstract or has too many complicated character names (foreign names translated into Chinese will be very long and hard for Chinese to remember). Nonetheless, we emphasize to annotators to mainly read the snippet to avoid ChatGPT's error. As for the possible choices, the annotators can choose \textit{Definitely Is}, \textit{Presumably Is}, and \textit{Is Not}, in which \textit{Definitely Is} and \textit{Presumably Is} stand for ``YES'' and \textit{Is Not} stands for ``NO''. \textit{Presumably Is} snippets are those recap snippets not that strongly confirmed by annotators but still plot-related and causal-related (usually those snippets not in the main plot but in sub-plots and sub-arcs). 

\clearpage
\clearpage
\section{Recaps Obtained from Events for TV Productions}\label{event_recap_appendix}

\begin{figure*}
    \centering
    \scalebox{0.46}{
    \includegraphics{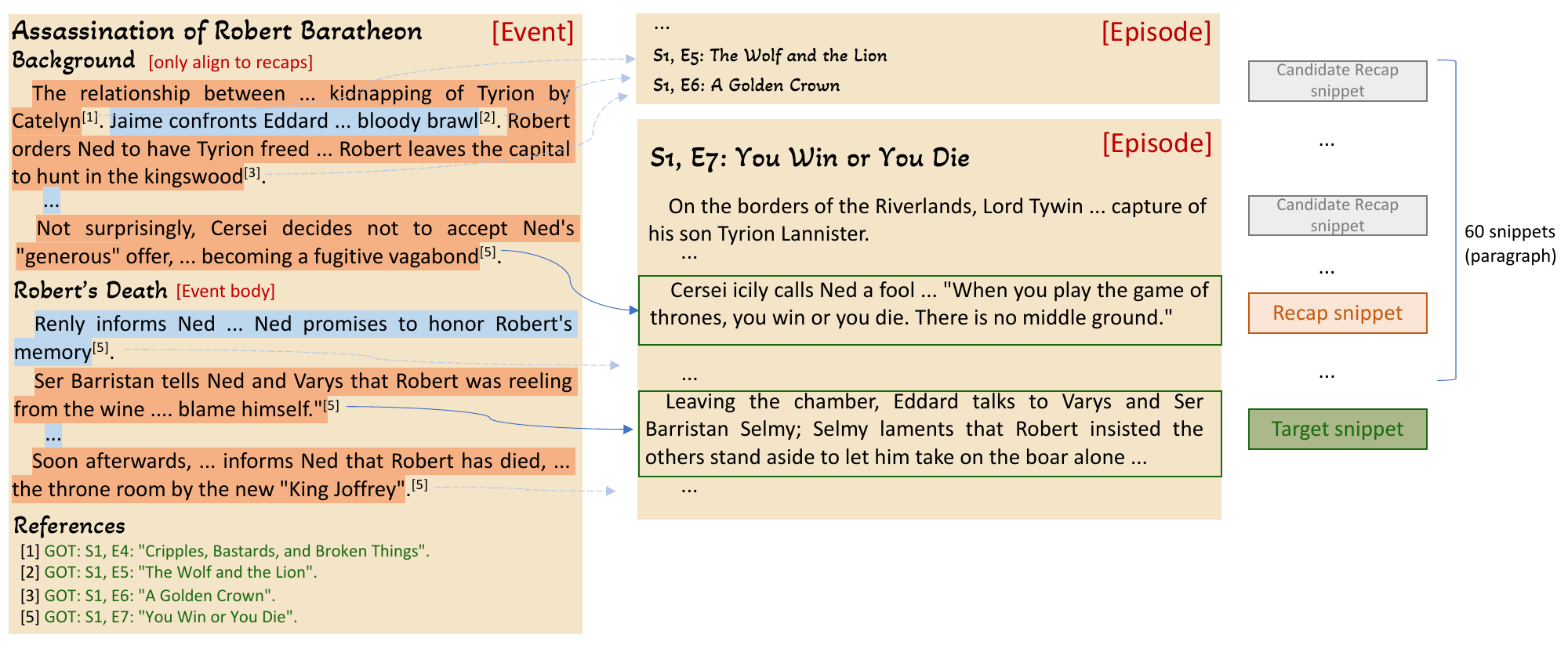}
    }
    \caption{The process of automatic recap snippet label construction for GOT and AOT. We utilize the episode reference citing paragraphs or sentence groups to align with the paragraphs of the cited episodes. }
    \label{event_align}
\end{figure*}

The process of automatic recap snippet label construction is illustrated in Fig.~\ref{event_align}. 

We mentioned we utilize SBERT to simply align event paragraphs with synopsis paragraphs instead of using vecalign and GNAT. The reason for this is that vecalign is not good at the imbalanced cases where a paragraph needs to find its corresponding paragraphs from a list of paragraphs (1-to-n alignment, vecalign is good at n-to-n alignment). GNAT is said to be good at imbalanced alignment, but we tried it and found that it is even worse than vecalign. 

To make the process of getting recaps from events clearer, we answer several questions below:

\textit{Q1: How many events are involved in GOT and AOT?}

We get 66 events from \textit{Game of Thrones} and 11 events from \textit{Attack on Titan}. The detailed event names can be viewed in Tab.~\ref{event_list}. 

\textit{Q2: Are there any synopsis paragraphs acting as target snippets in several events?}

The answer is yes. There are some target snippets in GOT and AOT appearing in more than one event. 13 out of 204 target snippets appear in more than one event in GOT. 11 out of 194 target snippets appear in more than one event in AOT. To this end, all of our experiments with GOT and AOT add the event name to remind models of which event the target snippet is at. As for the ablation study on the event name in Section 6.2 \textit{The Effect of Event Names}, we present the results only on those target snippets belonging to only one event (191/204 for GOT and 183/194 for AOT). 

\begin{table*}[]
\centering
\scalebox{0.6}{
\begin{tabular}{@{}l|ccc|ccc|ccc|ccc|ccc|c@{}}
\toprule\toprule
\multirow{2}{*}{}       & \multicolumn{3}{c|}{Notre-Dame de Paris} & \multicolumn{3}{c|}{Demi-Gods \& Semi-Devils} & \multicolumn{3}{c|}{The Count of Monte Cristo}  & \multicolumn{3}{c|}{Game of Thrones} & \multicolumn{3}{c|}{Attack on Titan} & \multicolumn{1}{c}{Avg.} \\ \cmidrule(l){2-17} 
                                 & R       & P       & F1       & R~~        & P       & F1     & R~~     & P     & F1   & R     & P     & F1    & R     & P     & F1  & F1  \\ \midrule\midrule
\multicolumn{17}{c}{\textit{zero-shot}} \\

Select-All       & 100.0   & 9.39   & 17.17     & 100.0~~  &  12.75  & 22.62  & 100.0~~& 20.44 & 33.94 & 100.0 & 8.18 & 15.12 & 100.0  & 18.47  & 31.18 & 24.01 \\ 
ChatGPT Listwise        & 15.78   & 10.66   & 12.72     & 15.06~~  &  13.43  & 14.20  & 24.56~~& 22.61 & 23.55 & 34.14 & 29.81 & 31.83 & 25.55  & 27.59  & 26.53 & 21.77 \\ 
\quad  + Char-Filter    & 20.74   & 21.05   & 20.89    & 20.67~~  &  16.76  & 18.51  & 30.29~~& 43.82 & 35.82 & 34.30 & 39.82 & 39.82 & 22.08  & 34.20  & 26.83 & 28.37 \\
\midrule
\multicolumn{17}{c}{\textit{fine-tune}} \\
Roberta Pw                 & 33.78   & 16.02   & 21.73    & 39.04    &  18.18  & 24.81  & 27.74 & 31.43 & 29.47& 52.22 & 16.11 & 24.62 & 22.46 & 36.83 & 27.90 & 25.71 \\ 
SBERT Pw                   & 54.98   & 12.90   & 20.90    & 76.73    &  13.66  & 23.19  & 51.85 & 22.39 & 31.27& 42.67 & 18.17 & 25.49 & 37.55 & 32.33 & 34.75 & 27.12 \\ 
Llama2Lora Pw              & 44.19   & 28.43   & 34.60    & 51.59    &  33.39  & 40.54  & 45.27 & 59.52 & 51.43& 49.68 & 43.10 & 46.16 & 55.52    & 34.57 & 42.61 & 43.07 \\ 
\bottomrule\bottomrule
\end{tabular}
}
\caption{Recall, Precision, F1 scores of different methods on the English evaluation data of RECIDENT. }
\label{english_appendix}
\end{table*}

\section{The full Recall, Precision, F1 scores}\label{free_f1}

The Recall, Precision, and F1 scores on English and Chinese evaluation data are shown in Tab.~\ref{english_appendix} and Tab.~\ref{chinese_appendix}. When predicting the recap snippets freely, ChatGPT can achieve better results compared to picking up the top 5 recap snippets, which indicates that ChatGPT is not good at being a ranker for recap identification. As seen from these tables, models tend to achieve higher Recall scores than Precision scores, which means that models prefer to predict a lot of candidate snippets as recaps. It may bring the problem of raising the number of predicted recaps to achieve a higher Recall score to reconcile the F1 score. 

From the tables, we also can see that fine-tuned LLMs can be better predictors than others, which aligns with our findings in Tab.~\ref{chinese_main} and~\ref{english_main}. Access to more training data or adapting unsupervised training on LLMs can be a possible solution. 

\begin{table*}[]
\centering
\scalebox{0.8}{
\begin{tabular}{@{}l|ccc|ccc|ccc|c@{}}
\toprule\toprule
\multirow{2}{*}{}       & \multicolumn{3}{c|}{Notre-Dame de Paris} & \multicolumn{3}{c|}{Demi-Gods \& Semi-Devils} & \multicolumn{3}{c|}{The Count of Monte Cristo} & \multicolumn{1}{c}{Avg.}\\ \cmidrule(l){2-11} 
                        & R            & P              & F1        & R~~~      & P            & F1       & R~~~      & P            & F1   & F1 \\ \midrule\midrule
\multicolumn{11}{c}{\textit{zero-shot}} \\

Select-All     & 100.0        & 9.39           & 17.17    & 100.0~~~   &  12.75       & 22.62   & 100.0~~~  &  20.44       & 33.94 & 24.58 \\
ChatGPT Listwise     & 15.11       & 9.31           & 11.52    & 15.13~~~   &  20.79       & 17.52   & 24.90~~~  &  22.48       & 23.63 & 17.56 \\
\quad  + Char-Filter    & 21.26       & 24.41          & 22.73    & 26.35~~~   &  18.64       & 21.88   & 25.88~~~  &  37.62       & 30.67 & 25.09 \\
\midrule
\multicolumn{11}{c}{\textit{fine-tune}} \\
Roberta Pw              &  48.39       &  17.48         &  25.68   & 42.28     &   20.07      &  27.22 & 29.83~~~  &  33.72       & 31.66 & 28.19 \\
SBERT Pw                &  51.88       &  12.60         &  20.28   & 53.94     &   15.20      &  24.20 & 58.62~~~  &  22.19       & 32.19 & 25.56 \\
InternLM2 Pw            &  66.83       &  22.61         &  33.79   & 44.77     &   39.52      &  41.98 & 55.53~~~  &  59.96       & 57.66 & 44.48 \\
\midrule
Human Performance$^{*}$ & 67.62        & 51.67          & 58.13    & 58.56     & 64.10        & 60.82  &  85.39  &   65.10    &   73.88      &  64.38 \\
\bottomrule\bottomrule
\end{tabular}
}
\caption{Recall, Precision, F1 scores of different methods on the Chinese evaluation data of RECIDENT.}
\label{chinese_appendix}
\end{table*}

\section{The Training of Line2Note}

\begin{table}[]
\centering
\scalebox{0.7}{
\begin{tabular}{cl|cc|cc}
\toprule
\multicolumn{2}{c|}{\multirow{2}{*}{Method}}                       & \multicolumn{2}{c|}{Dev} & \multicolumn{2}{c}{Test} \\ \cline{3-6} 
\multicolumn{2}{c|}{}                                              & ACC         & HIT@1      & ACC         & HIT@1      \\ \midrule

\multicolumn{1}{c|}{\multirow{4}{*}{l2n}} & RoBERTa vanilla        & 8.66        & 10.54      & 8.93        & 11.61      \\
\multicolumn{1}{c|}{}                     & SBERT vanilla          & 8.62        & 10.18      & 9.10        & 11.39      \\
\cline{2-6}
\multicolumn{1}{c|}{}                     & RoBERTa                & 47.76       & 57.99      & 44.71       & 57.49      \\
\multicolumn{1}{c|}{}                     & SBERT                  & 43.53       & 53.05      & 39.47       & 50.90      \\
\bottomrule
\end{tabular}
}
\caption{Accuracy scores and HIT@1 scores of PLMs and them further unsupervised Line2Note fine-tuned on our collected reader notes. }
\label{l2n_pretrain}
\end{table}

The accuracy scores and HIT@1 scores of untrained RoBERTa, SBERT, and fine-tuned RoBERTa, SBERT are shown in Tab.~\ref{l2n_pretrain}. RoBERTa can achieve better performance than SBERT after Line@Note training, which can further indicate that SBERT's ability of sentence similarity computing is hurt by plot-related fine-tuning. This also indicates that snippets' plot association is more than just similarity computing. 

\section{More Illustrations of the Effect of Distance}\label{more_dis}

\begin{figure*}
    \centering
    \scalebox{0.5}{
    \includegraphics{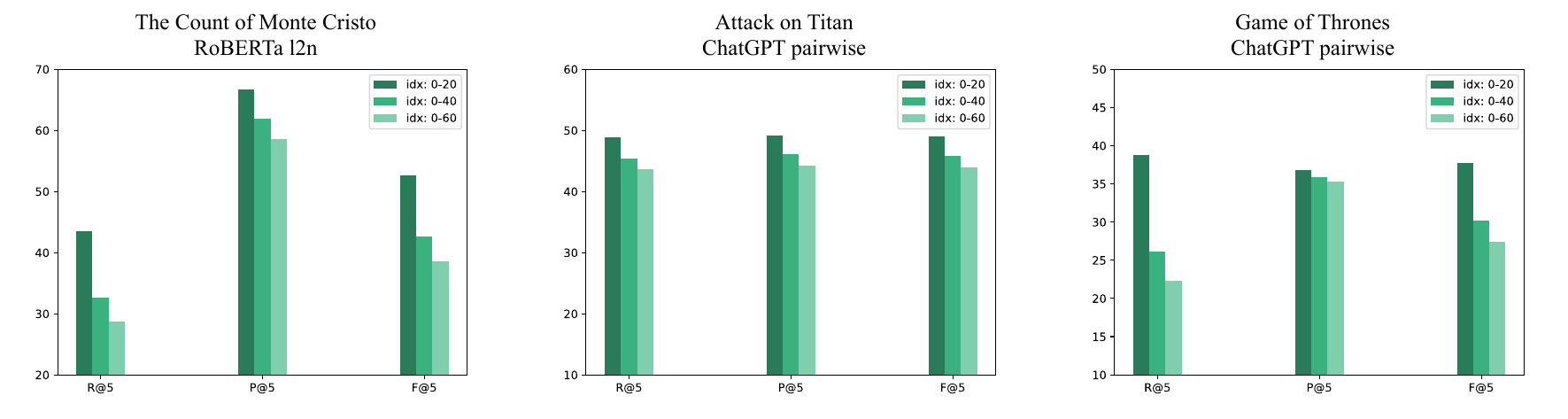}
    }    
    \caption{The performance of SBERT l2n and ChatGPT pairwise when only considering the nearest 20/40/60 snippets on TCOMC, GOT, and AOT. }
    \label{dis_performance_appendix}
\end{figure*}

The performance of models on TCOMC, GOT, and AOT with more snippets considered as candidate snippets along the increasing distance is shown in Fig.~\ref{dis_performance_appendix}.

\begin{table}[t!]
\centering
\scalebox{0.5}{
\begin{tabular}{ll}
\toprule
Game of Thrones & Attack on Titan \\ \midrule
\textit{Assassination of Jon Arryn} & \textit{Fall of Wall Maria (Anime)} \\
\textit{Assassination of Robert Baratheon} & \textit{Struggle for Trost} \\
\textit{Battle in the hills below the Golden Tooth} & \textit{Assault on Stohess} \\
\textit{Battle of Whispering Wood} & \textit{Wall Rose invasion (Anime)} \\
\textit{Battle on the Green Fork} & \textit{57th Exterior Scouting Mission (Anime)} \\
\textit{Execution of Eddard Stark} & \textit{Coup d'état (Anime)} \\
\textit{First trial by combat of Tyrion Lannister} & \textit{Great Titan War (Anime)} \\
\textit{Massacre in the Red Keep} & \textit{Marley Mid-East War (Anime)} \\
\textit{Skirmish at Littlefinger's brothel} & \textit{Raid on Liberio (Anime)} \\
\textit{Assassination of Renly Baratheon} & \textit{Battle of Shiganshina District (Anime)} \\
\textit{Attack on Deepwood Motte} & \textit{Rumbling (Anime)} \\
\textit{Battle at the Yellow Fork} & \textit{  } \\
\textit{Battle of the Blackwater} & \textit{  } \\
\textit{Battle of the Fords} & \textit{  } \\
\textit{Capture of Winterfell} & \textit{  } \\
\textit{Fight at the Fist} & \textit{  } \\
\textit{Fight at the holdfast} & \textit{  } \\
\textit{Massacre in King's Landing} & \textit{  } \\
\textit{Riots in King's Landing} & \textit{  } \\
\textit{Sack of Winterfell} & \textit{  } \\
\textit{Surrender of the Crag} & \textit{  } \\
\textit{Taking of Ashemark} & \textit{  } \\
\textit{Battle in Yunkai} & \textit{  } \\
\textit{Fall of Astapor} & \textit{  } \\
\textit{Fall of Harrenhal} & \textit{  } \\
\textit{Leech ritual} & \textit{  } \\
\textit{Mutiny at Craster's Keep} & \textit{  } \\
\textit{Red Wedding} & \textit{  } \\
\textit{Assassinations in the Tower of the Hand} & \textit{  } \\
\textit{Battle for the Wall} & \textit{  } \\
\textit{Brawl in a tavern in the Riverlands} & \textit{  } \\
\textit{Purple Wedding} & \textit{  } \\
\textit{Raid on Craster's Keep} & \textit{  } \\
\textit{Sack of Mole's Town} & \textit{  } \\
\textit{Second trial by combat of Tyrion Lannister} & \textit{  } \\
\textit{Siege of Moat Cailin} & \textit{  } \\
\textit{Assassination of Myrcella Baratheon} & \textit{  } \\
\textit{Confrontation in the Water Gardens} & \textit{  } \\
\textit{Mutiny at Castle Black} & \textit{  } \\
\textit{Assassination of Balon Greyjoy} & \textit{  } \\
\textit{Assassinations at the Twins} & \textit{  } \\
\textit{Assassinations at Winterfell} & \textit{  } \\
\textit{Battle at the cave of the Three-Eyed Raven} & \textit{  } \\
\textit{Battle of the Bastards} & \textit{  } \\
\textit{Coup in Dorne} & \textit{  } \\
\textit{Fight by Deepwood Motte} & \textit{  } \\
\textit{Massacre of the Khalar Vezhven} & \textit{  } \\
\textit{Second siege of Riverrun} & \textit{  } \\
\textit{Siege of Astapor} & \textit{  } \\
\textit{Standoff at the Great Sept of Baelor} & \textit{  } \\
\textit{Taking of Riverrun} & \textit{  } \\
\textit{Assault on the Targaryen fleet} & \textit{  } \\
\textit{Battle of the Goldroad} & \textit{  } \\
\textit{Breaching of the Wall} & \textit{  } \\
\textit{Dragonpit Summit} & \textit{  } \\
\textit{Fall of Casterly Rock} & \textit{  } \\
\textit{Sack of Highgarden} & \textit{  } \\
\textit{Trial of Petyr Baelish} & \textit{  } \\
\textit{Wight hunt} & \textit{  } \\
\textit{Assassination of Daenerys Targaryen} & \textit{  } \\
\textit{Battle at Dragonstone} & \textit{  } \\
\textit{Battle of King's Landing} & \textit{  } \\
\textit{Battle of Winterfell} & \textit{  } \\
\textit{Fall of Last Hearth} & \textit{  } \\
\textit{Rescue of Yara Greyjoy} & \textit{  } \\
\textit{Retaking of the Iron Islands} & \textit{  } \\ \bottomrule
\end{tabular}
}
\caption{Event name lists of GOT and AOT. }
\label{event_list}
\end{table}

\end{document}